\documentclass{article}

\usepackage{microtype}
\usepackage{graphicx}
\usepackage{booktabs} 

\usepackage{hyperref}

\usepackage[accepted]{icml2025}

\usepackage{amsmath}
\usepackage{amssymb}
\usepackage{mathtools}
\usepackage{amsthm}

\usepackage[capitalize,noabbrev]{cleveref}

\theoremstyle{plain}

\theoremstyle{definition}

\theoremstyle{remark}

\usepackage[textsize=tiny]{todonotes}

\usepackage{url}
\usepackage{hyperref}

\usepackage{graphicx}
\usepackage{amsfonts}
\usepackage{amsmath}
\usepackage{siunitx}
\usepackage{amssymb}
\usepackage{pifont}
\usepackage{nicefrac}
\usepackage{multirow}
\usepackage[font=small]{caption}
\usepackage{soul}
\usepackage{booktabs}
\usepackage{xcolor}
\usepackage{comment}
\usepackage{sidecap}
\usepackage{xspace}
\usepackage[most]{tcolorbox}
\usepackage{enumitem}
\usepackage{ulem}
\usepackage{makecell}
\usepackage{subcaption}

\newsavebox{\tablebox}

\usepackage[utf8]{inputenc}
\usepackage[T1]{fontenc}
\usepackage{CJKutf8}

\definecolor{bblue}{HTML}{4F81BD}
\definecolor{rred}{HTML}{c4260b}
\definecolor{ggreen}{HTML}{098c1f}
\definecolor{ppurple}{HTML}{9F4C7C}
\definecolor{oorange}{HTML}{F79646}

\usepackage{tablefootnote}

\icmltitlerunning{Encoder-Decoder Gemma: Improving the Quality-Efficiency Trade-Off via Adaptation}

\begin{document}

\twocolumn[
\icmltitle{Encoder-Decoder Gemma: Improving the Quality-Efficiency Trade-Off via Adaptation}



\icmlsetsymbol{core}{*}

\begin{icmlauthorlist}

\icmlauthor{Biao Zhang}{core}
\icmlauthor{Fedor Moiseev}{core}
\icmlauthor{Joshua Ainslie}{core}
\icmlauthor{Paul Suganthan}{core}
\icmlauthor{Min Ma}{core}
\icmlauthor{Surya Bhupatiraju}{}
\icmlauthor{Fede Lebron}{}
\icmlauthor{Orhan Firat}{}
\icmlauthor{Armand Joulin}{}
\icmlauthor{Zhe Dong}{core}
\end{icmlauthorlist}


\icmlcorrespondingauthor{Biao Zhang}{biaojiaxing@google.com}
\icmlcorrespondingauthor{Zhe Dong}{zhedong@google.com}


\vskip 0.3in
]



\printAffiliationsAndNotice{\textsuperscript{*} Core Contributor. Google}  

\begin{abstract}

While decoder-only large language models (LLMs) have shown impressive results, encoder-decoder models are still widely adopted in real-world applications for their inference efficiency and richer encoder representation. In this paper, we study a novel problem: adapting pretrained decoder-only LLMs to encoder-decoder, with the goal of leveraging the strengths of both approaches to achieve a more favorable quality-efficiency trade-off. We argue that adaptation not only enables inheriting the capability of decoder-only LLMs but also reduces the demand for computation compared to pretraining from scratch. We rigorously explore different pretraining objectives and parameter initialization/optimization techniques. Through extensive experiments based on Gemma 2 (2B and 9B) and a suite of newly pretrained mT5-sized models (up to 1.6B), we demonstrate the effectiveness of adaptation and the advantage of encoder-decoder LLMs. Under similar inference budget, encoder-decoder LLMs achieve comparable (often better) pretraining performance but substantially better finetuning performance than their decoder-only counterpart. For example, Gemma 2B-2B outperforms Gemma 2B by $\sim$7\% after instruction tuning. Encoder-decoder adaptation also allows for flexible combination of different-sized models, where Gemma 9B-2B significantly surpasses Gemma 2B-2B by $>$3\%. The adapted encoder representation also yields better results on SuperGLUE.
We will release our checkpoints to facilitate future research.

\end{abstract}

\section{Introduction}\label{sec:introduction}

Neural network architectures are often designed to incorporate certain assumptions or inductive biases regarding the input data, leading to either improved model performance or better computational efficiency, if not both. Unlike the popular decoder-only architecture used for large language model (LLM)~\cite{brown2020language}, the encoder-decoder architecture adopts separate modeling modules -- an encoder for input understanding and a decoder for output generation~\cite{transformer}. This separation decouples parameters for different functionalities and thus enjoys higher freedom in handling contextual representation and challenging tasks~\cite{tay2022ul2,pmlr-v162-wang22u}. It also offers high flexibility in changing the encoder and decoder size (e.g., a large encoder paired with a small decoder) to control the quality-efficiency trade-off~\cite{kasai2020deep,pmlr-v162-zhang22h}, an increasingly important aspect for LLM deployment~\cite{team2024gemini}. Despite these benefits, however, the study on encoder-decoder LLMs receive little to no attention nowadays.

In this paper, we revisit this classical architecture by exploring the following question: can we get strong(er) encoder-decoder LLMs by adapting from existing pretrained decoder-only LLMs? We consider the adaptation more significantly than pretraining new models from scratch since pretraining is resource-intensive and powerful decoder-only models at different sizes are already widely available~\cite{dubey2024llama,team2024gemma,liu2024deepseek,yang2024qwen2,jiang2024mixtral}. Our hypothesis is that, by reusing parameters from decoder-only models, we can accelerate training and effectively transfer their internal knowledge to encoder-decoder, preserving (even enhancing) their capabilities. Note adaptation also allows for pairing varying-sized decoder-only models to achieve specific quality-efficiency considerations. Yet, the optimal method for such adaptation and the extent to which performance can be improved remain open questions, which we aim to address rigorously.

\begin{figure*}[t]
\centering
\small
\includegraphics[width=0.7\textwidth]{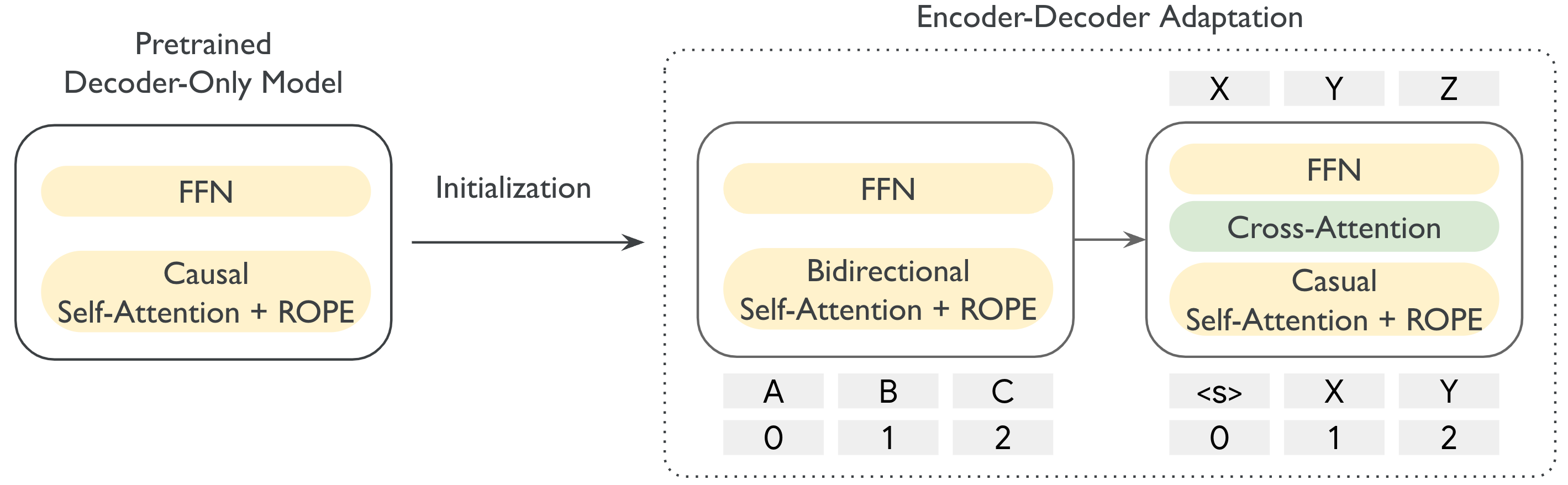}
\caption{\label{fig:overview} Overview of our approach. We build encoder-decoder models by adapting from pretrained decoder-only models. Model architecture and parameters are inherited from the decoder-only model except the cross-attention, for which we adopt different initialization methods depending on the encoder and decoder size. ``ROPE'': rotary embedding; ``FFN'': feed-forward layer.}
\end{figure*}

We employ Gemma 2~\cite{team2024gemma} as the testbed. As shown in Figure \ref{fig:overview}, the encoder-decoder architecture follows the original Transformer~\cite{transformer} but equipped with Gemma 2 modifications. The key idea behind the adaptation is to initialize the parameters of the encoder-decoder model from pretrained decoder-only model(s) as a warmup and then pretrain or adapt all parameters with self-supervised learning. Depending on whether the encoder and the decoder share the same configuration, we propose different initialization and optimization strategies for the cross-attention layer. We also compare different pretraining objectives, including prefix language modeling with knowledge distillation~\cite{hinton2015distilling} and UL2~\cite{tay2022ul2}. Apart from Gemma 2 2B and 9B, we pretrain a series of small models to better understand the adaptation at different scales.

To thoroughly evaluate model performance, we adopt different benchmarks for pretrained and instruction-tuned models respectively, each covering a range of established academic evaluations. In addition, we use SuperGLUE~\cite{wang2019superglue} to measure the quality of the learned contextual representations. Our main findings are below:
\begin{itemize}
    \item Leveraging pretrained decoder-only LLMs is an effective way to build powerful encoder-decoder LLMs, which yields substantially improved downstream performance particularly after instruction tuning under similar inference flops.
    \item Our adaptation method is highly flexible, allowing for pairing large encoder with small decoder, such as 9B-2B, with significant quality gains over Gemma 2 2B but similar generation latency.
    \item Adaptation is not only more compute efficient but also more effective than pretraining from scratch.
    \item Pretraining objective matters. Models trained with prefix language modeling and knowledge distillation are generally better at generative tasks, while UL2 models have better encoder representations.
\end{itemize}

\section{Related Work}

While the decoder-only architecture has become the \textit{de facto} standard for LLMs, the debate between encoder-decoder and decoder-only modeling is still not conclusive. Many prior studies proposed different approaches to pretrain strong encoder-decoder models, e.g., MASS~\cite{song2019mass}, T5~\cite{t5_paper}, mT5~\cite{xue-etal-2021-mt5}, byT5~\cite{xue-etal-2022-byt5}, BART~\cite{lewis-etal-2020-bart}, and OpenBA~\cite{li2023openba}. \citet{tay2022ul2} compared different pretraining objectives, highlighting the superiority of UL2 and encoder-decoder modeling. \citet{pmlr-v162-zhang22h} systematically examined the scaling behavior of both architectures on machine translation, showing their similarity when adequate objectives are applied. \citet{pmlr-v162-wang22u} thoroughly explored different modeling choices and training objectives with a focus on LLM zero-shot generalization. They discovered that encoder-decoder LLMs after instruction tuning achieve the best performance, echoing with our experiments. They also studied adaptation, but it is between different pretraining objectives rather than from decoder-only LLMs to encoder-decoder LLMs.

Leveraging pretrained models for encoder-decoder modeling has been extensively explored. In the BERT era~\cite{devlin-etal-2019-bert}, researchers developed different ways of utilizing it to enhance encoder-decoder performance on downstream tasks, such as machine translation~\cite{zhu2020incorporating,clinchant-etal-2019-use,yang2020towards}, grammatical error correction~\cite{kaneko2020encoder}, summarization~\cite{liu2019text}, and text generation~\cite{chen2019distilling}. Our work follows a similar spirit but is based on pretrained decoder-only LLMs and focuses on developing general-purpose encoder-decoder LLMs.

Another related direction is the development of inference friendly LLMs. Techniques for improving inference efficiency are many, ranging from quantization~\cite{dettmers2023case}, key-value cache optimization~\cite{corallo-papotti-2024-finch}, recurrent modeling~\cite{gu2023mamba,botev2024recurrentgemma}, to strong small LLMs with improved pretraining~\cite{abdin2024phi,liu2024mobilellm}, to name a few. While these techniques offer significant efficiency gains, their focus is fundamentally distinct and complementary to our proposed encoder-decoder adaptation, i.e., both approaches can be used in conjunction to realize greater overall efficiency.

\section{Approach: Encoder-Decoder Adaptation}

\begin{table*}[t]
\centering
\centering
\small
\begin{tabular}{lrrrrrrr}
\toprule
& \multirow{2}{*}{\#Layers} & \multirow{2}{*}{$d_{model}$} & \multirow{2}{*}{$d_{ffn}$} & \multirow{2}{*}{\#heads (q/kv)} & \multirow{2}{*}{$d_{head}$} & \multicolumn{2}{c}{\#Params} \\
& & & & & & Decoder-Only & Encoder-Decoder \\
\midrule
2B & 26 & 2304 & 18432 & 8/4 & 256 & 2.0B & 4.0B (2B-2B) \\
9B & 42 & 3584 & 28672 & 16/8 & 256 & 8.3B & 16.7B (9B-9B) \\
\midrule
S (Small) & 8 & 512 & 1024 & 8/8 & 64 & 14.7M & 29.4M (S-S) \\
B (Base) & 12 & 768 & 2048 & 12/12 & 64 & 56.7M & 113.3M (B-B) \\
L (Large) & 24 & 1024 & 2816 & 16/16 & 64 & 204.6M & 409.1M (L-L) \\
XL (Xlarge) & 24 & 2048 & 5120 & 32/32 & 64 & 780.3M & 1.6B (XL-XL) \\
\bottomrule
\end{tabular}
\caption{\label{tab:model_config} Model configurations. \textit{\#Layers}: number of layers; $d_{model/ffn/head}$: model/feed-forward/head dimension; \textit{\#heads (q/kv)}: number of query/key-value heads. \textit{\#Params}: number of model parameters excluding embeddings. For encoder-decoder models, we show the number of parameters for the balanced architecture, e.g. 2B-2B. The 9B-2B model has 10.4B parameters. ``B/M'': billion/million.}
\end{table*}

\subsection{Architecture}

Pretraining LLMs is both compute and time intensive. To reduce the amount of training required, we propose to adapt existing decoder-only LLMs to encoder-decoder and leverage pretrained decoder-only checkpoints for initialization, as shown in Figure \ref{fig:overview}. Due to this, we keep the encoder-decoder architecture as similar as possible to original decoder-only model, only introducing changes when necessary. This results in the following architecture:
\begin{enumerate}
\item \textbf{Encoder} has exactly the same architecture as the decoder-only model, but self-attention is switched from causal to bidirectional. We provide ablations in Section \ref{sec:discussion} that illustrate the critical effect of bidirectional attention on downstream performance.
\item In each \textbf{Decoder} block, FFN and self-attention parts are identical to the corresponding parts in decoder-only models, and cross-attention has the same number of heads and head dimension as self-attention, but attends to the whole output of the encoder.
\end{enumerate}

We base our study on Gemma 2~\cite{team2024gemma}. But note our approach is highly flexible and isn't restricted to specific decoder-only architectures. We can easily apply our method to other model families, such as LLaMA~\cite{dubey2024llama}, QWen~\cite{yang2024qwen2}, and DeepSeek~\cite{liu2024deepseek}. In theory, we can also adapt decoder-only models from different families, such as pairing LLaMA models with QWen models.

In addition, our approach allows for unbalanced encoder-decoder models, where the decoder is significantly smaller than the encoder. This provides better support for applications where input processing capabilities are more important than generative capacity. For example, for summarization, deep understanding of the input text is often more important than the generation part, as it doesn't need to generate any new information. As a result, generation time is significantly reduced, while providing competitive quality. 

\subsection{Initialization}

When initializing an encoder-decoder model from a decoder-only checkpoint, we try to map every layer to the most similar weight in the decoder-only checkpoint. In particular, the encoder is fully initialized from the decoder-only checkpoint, as it doesn't introduce any new weights. In the decoder, FFN and self-attention subblocks are initialized from the FFN and self-attention weights from the corresponding layers in the decoder-only checkpoint.

Cross-attention is initialized from self-attention weights in the balanced setup where encoder and decoder have the same configuration. Otherwise, we first initialize cross-attention from scratch and then finetune it for the first $K$ steps as a warmup while freezing other model parameters. After $K$ steps, all model parameters are tuned.

\subsection{Pretraining Objective}

Decoder-only pretraining often adopts causal language modeling on a single sequence. In contrast, encoder-decoder adaptation requires separate input and target sequences to be fed to the encoder and decoder separately. We explore two classical pretraining objectives for encoder-decoder modeling: prefix language modeling (PrefixLM) and UL2~\cite{tay2022ul2,pmlr-v162-wang22u}.

PrefixLM behaves similar to causal language modeling except for its prefix condition. To simplify the preprocessing, we split a sequence equally into two halves, the first half used as input and the second one as target. This also eases the adoption of knowledge distillation from decoder-only models. UL2 is more complicated. It is composed of several denoising tasks at different levels of complexity. We prepare UL2 data following~\citet{tay2022ul2}. We compare their performance in experiments.

\section{Setup}

\paragraph{Data Setting} Our data for pretraining and instruction tuning -- including supervised finetuning (SFT) and reinforcement learning from human feedback (RLHF) -- follow Gemma 2~\cite{team2024gemma}. For the adaptation, we preprocess the Gemma 2 pretraining data (8 trillion tokens) with PrefixLM and UL2. Note Gemma 2 pretraining data comes with knowledge distillation. We preserve this information for PrefixLM while adopting ground-truth targets for UL2 as mapping the teacher logits to UL2 is non-trivial. The preprocessed data has an input-output sequence length of 4096-4096 and 8192-8192 for PrefixLM and UL2, respectively. We adapt our models on up to 2 trillion tokens.

\paragraph{Model Setting} We use Gemma 2 (2B and 9B) as the base decoder-only LLM. We also pretrain several smaller models (Small, Base, Large, and XL) following mT5 configurations~\cite{xue-etal-2021-mt5} under the Gemma 2 framework, and then adapt them to encoder-decoder LLMs. Detailed model configurations are given in Table \ref{tab:model_config}.

\paragraph{Evaluation} We employ diverse academic evaluation datasets to evaluate different capabilities of LLMs. Concretely, we use the following benchmarks:
\begin{itemize}
    \item Pretraining (PT) benchmark: Boolq~\cite{clark2019boolq}, SIQA~\cite{sap2019socialiqa}, PIQA~\cite{bisk2020piqa}, ARC-c\&ARC-e~\cite{allenai:arc}, MMLU~\cite{hendrycks2021measuring}, MMLU Pro~\cite{wang2024mmlu}, HellaSwag~\cite{zellers2019hellaswag}, Winogrande~\cite{sakaguchi2021winogrande}, TruthfulQA~\cite{lin2021truthfulqa}, AGIEval~\cite{zhong2023agieval}, BBH~\cite{suzgun2022challenging}, DROP~\cite{dua2019drop}, GPQA~\cite{rein2023gpqa}, GSM8K~\cite{cobbe2021gsm8k}, HumanEval~\cite{chen2021evaluating}, Lambada~\cite{paperno2016lambada}, MATH-500~\cite{hendrycks2021measuring}, MBPP~\cite{austin2021program}, NQ~\cite{kwiatkowski-etal-2019-natural}, TriviaQA~\cite{2017arXivtriviaqa}, and WMT23~\cite{kocmi-etal-2023-findings}. We perform zero/few-shot prompting for pretrained LLMs, and report the averaged result as \textit{PT score}.
    \item Instruction-tuning (IT) benchmark: GSM8K, MMLU, MMLU Pro, MBPP, HumanEval, MATH-500, BBH, GPQA (Diamond), WMT23, and MGSM~\cite{shi2022language}. We perform zero/few-shot prompting with task-specific instruction for instruction-tuned models, and report the averaged result as \textit{IT score}.
    \item SuperGLUE~\cite{10.5555/3454287.3454581}: we use this benchmark to examine the learned contextual representation. We stack a task-specific head on the representation of the last token in the encoder (decoder) of the encoder-decoder (decoder-only) LLM, and finetune all parameters on the training set. Learning rate, batch size, and dropout are grid-searched for each task. We reformulate all tasks as classification tasks and report averaged dev-set accuracy over COPA, WIC, WSC, RTE, MultiRC, CB, and Boolq.
\end{itemize}

For generative tasks, we always apply greedy sampling. We perform pretraining, SFT, and RLHF based on the Gemma 2 recipe except for the learning rate which we tune empirically for encoder-decoder LLMs. In unbalanced encoder-decoder adaptation, e.g. 9B-2B, we set the cross-attention warmup step $K$ to 1000.

\section{Results}

\begin{figure}[t]
\centering
\small

\includegraphics[width=0.33\textwidth]{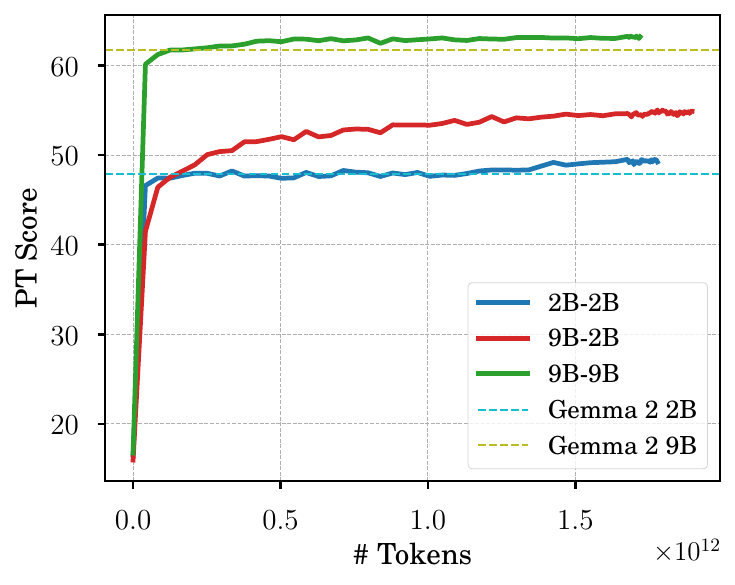}

\caption{\label{fig:pt_vs_step} Pretraining performance as a function of the number of pretrained tokens during the adaptation.}
\end{figure}

\paragraph{The encoder-decoder adaptation converges rapidly, particularly for balanced architectures.}

While adaptation leverages pretrained parameters for initialization, whether and how this benefits model convergence is still questionable. Figure \ref{fig:pt_vs_step} shows the change of PT performance with respect to the amount of pretrained tokens. Obviously, adaptation is very computationally efficient, converging quickly and achieving similar performance to its decoder-only counterpart after only tens of billions of tokens. Balanced architectures (2B-2B and 9B-9B) converge much faster than the unbalanced ones (9B-2B) since all parameters in the former are initialized from pretrained decoder-only models while the cross-attention in the latter is randomly initialized.

We also notice that additional pretraining improves balanced models a little on average but substantially benefits some tasks, like GSM8K and DROP. Besides, 9B-2B performance increases consistently during the adaptation, quickly surpassing Gemma 2 2B and moving towards Gemma 2 9B. This demonstrates the feasibility of encoder-decoder adaptation from varying-sized decoder-only LLMs, as well as its ability to utilize the knowledge from pretrained models.

\begin{table*}[t]
\centering
\centering
\small

\subcaptionbox{\label{tab:pt_it_performance} Results on PT and IT benchmarks.}{
\begin{tabular}{lrrrrrr}
\toprule
& \multicolumn{3}{c}{PT Score} & \multicolumn{3}{c}{IT Score} \\
\cmidrule(lr){2-4} \cmidrule(lr){5-7}
& Gemma 2 & + PrefixLM & + UL2 & Gemma 2 & + PrefixLM & + UL2 \\
\midrule
2B-2B & 47.9 & 49.7 & \textbf{50.1} & (39.0) & \textbf{46.4} (46.1) & 42.4 \\
9B-2B & - & \textbf{55.0} & 52.9 & - & 49.3 (\textbf{50.6}) & 45.7 \\
9B-9B & 61.7 & 63.1 & \textbf{63.9} & (59.6) & 62.9 (\textbf{64.5}) & 61.5 \\
\midrule
S-S & \textbf{23.4} & 22.8 & 23.1 & 6.2 & 9.8 & \textbf{10.7} \\
B-B & 26.7 & \textbf{26.9} & 26.0 & 9.8 & \textbf{12.9} & 11.1 \\
L-L & \textbf{32.3} & 31.6 & 30.9 & 12.9 & 17.5 & \textbf{18.9} \\
XL-XL & \textbf{39.7} & 39.5 & 38.5 & 23.5 & \textbf{30.7} & 29.2 \\
\bottomrule
\end{tabular}
}
\vspace{\baselineskip}

\subcaptionbox{\label{tab:superglue_performance} Finetuned performance on SuperGLUE.}{
\begin{tabular}{lrrrrrr}
\toprule
& \multicolumn{3}{c}{PT Models} & \multicolumn{3}{c}{IT Models} \\
\cmidrule(lr){2-4} \cmidrule(lr){5-7}
& Gemma 2 & + PrefixLM & + UL2 & Gemma 2 & + PrefixLM & + UL2 \\
\midrule
2B-2B & 75.5 & 88.1 & 88.1 & (86.2) & 88.3 (87.9) & \textbf{90.5} \\
9B-2B & - & 90.2 & 90.7 & - & 90.6 (90.3) & \textbf{91.3} \\
9B-9B & 82.5 & 91.4 & \textbf{91.8} & (89.8) & \textbf{91.8} (91.4) & 91.6 \\
\midrule
S-S & 67.6 & \textbf{69.8} & 69.6 & 67.6 & 68.8 & 69.4 \\
B-B & 68.6 & 71.2 & 71.5 & 68.7 & 72.3 & \textbf{73.6} \\
L-L & 68.4 & 78.7 & 79.7 & 68.8 & 78.1 & \textbf{80.3} \\
XL-XL & 70.7 & 84.4 & 85.4 & 69.2 & 85.7 & \textbf{87.0} \\
\bottomrule
\end{tabular}
}
\caption{\label{tab:model_performance} Main results on PT, IT, and SuperGLUE benchmarks. ``Gemma 2'': decoder-only models; ``+PrefixLM/UL2'': encoder-decoder models adapted via prefix language modeling (\textit{with knowledge distillation})/UL2. We put Gemma 2 results into the corresponding encoder-decoder rows to save space, e.g. 2B-2B for Gemma 2 means Gemma 2 2B. Numbers in parentheses are for RLHFed models. Best results are in \textbf{bold}. \textit{Note PT and IT scores are not directly comparable since they are averaged over different tasks.}}
\end{table*}

\begin{table*}[t!]
\centering
\centering
\small

\subcaptionbox{\label{tab:detailed_pt_results} Results for pretrained models.}{
\begin{tabular}{lcrrrrr}
\toprule
& & \multicolumn{2}{c}{Gemma 2} & \multicolumn{3}{c}{Encoder-Decoder Adaptation} \\
\cmidrule(lr){3-4} \cmidrule(lr){5-7}
Task & Metric & 2B & 9B & 2B-2B & 9B-2B & 9B-9B \\
\midrule
MMLU & 5-shot & 51.9 & 71.1 & 46.8 & 60.3 & \textbf{71.3} \\
ARC-C & 25-shot & 55.5 & \textbf{69.1} & 52.0 & 59.9 & 65.0 \\
GSM8K & 5-shot & 23.7 & 63.2 & 41.7 & 48.7 & \textbf{72.8} \\
AGIEval & 3-5-shot & 31.5 & \textbf{53.3} & 35.0 & 43.6 & 53.1 \\
DROP & 3-shot, F1 & 53.3 & 71.5 & 61.4 & 66.9 & \textbf{75.7} \\
BBH & 3-shot, CoT & 40.2 & 68.9 & 51.9 & 51.6 & \textbf{74.7} \\
Winogrande & 5-shot & 65.2 & 74.3 & 69.5 & 68.1 & \textbf{78.7} \\
HellaSwag & 10-shot & 72.9 & \textbf{81.8} & 74.9 & 75.7 & 81.0 \\
\midrule
MATH-500 & 4-shot & 17.2 & 33.4 & 24.2 & 23.6 & \textbf{37.8} \\
ARC-e & 0-shot & 81.0 & \textbf{88.3} & 77.1 & 82.9 & 85.3 \\
PIQA & 0-shot & 78.4 & \textbf{81.6} & 79.0 & 78.3 & 81.1 \\
SIQA & 0-shot & 51.7 & \textbf{53.6} & 50.1 & 50.1 & 50.5 \\
Boolq & 0-shot & 75.5 & 77.5 & 75.6 & 84.6 & \textbf{85.6} \\
TriviaQA & 5-shot & 60.1 & \textbf{76.6} & 51.2 & 66.2 & 75.2 \\
NQ & 5-shot & 30.7 & \textbf{43.9} & 28.4 & 37.1 & 43.1 \\
HumanEval & pass@1 & 19.5 & 39.0 & 27.4 & 33.5 & \textbf{40.2} \\
MBPP & 3-shot & 30.4 & 52.0 & 37.4 & 43.4 & \textbf{55.6} \\
\midrule
Average & & 49.3 & 64.7 & 52.0 & 57.3 & \textbf{66.3} \\
\bottomrule
\end{tabular}
}
\vspace{\baselineskip}

\subcaptionbox{\label{tab:detailed_it_results} Results for RLHFed models.}{
\begin{tabular}{lcrrrrr}
\toprule
& & \multicolumn{2}{c}{Gemma 2} & \multicolumn{3}{c}{Encoder-Decoder Adaptation} \\
\cmidrule(lr){3-4} \cmidrule(lr){5-7}
Task & Metric & 2B & 9B & 2B-2B & 9B-2B & 9B-9B \\
\midrule
GSM8K & 11-shot & 58.0 & 84.3 & 70.7 & 73.8 & \textbf{88.6} \\
MMLU & 5-shot & 49.8 & 71.8 & 61.5 & 66.7 & \textbf{76.7} \\
MMLU Pro & 5-shot & 27.4 & 49.9 & 36.6 & 43.0 & \textbf{55.7} \\
MBPP & 3-shot & 37.8 & 59.2 & 44.0 & 49.8 & \textbf{64.8} \\
HumanEval & pass@1 & 43.3 & 65.9 & 47.6 & 55.5 & \textbf{72.0} \\
MATH-500 & 0-shot & 24.4 & 45.8 & 28.2 & 30.0 & \textbf{47.2} \\
BBH & 3-shot & 44.8 & 72.0 & 57.5 & 57.6 & \textbf{76.4} \\
GPQA & 0-shot & 24.8 & 29.9 & 27.5 & 32.6 & \textbf{35.7} \\
GPQA Diamond & 0-shot & 27.8 & 29.8 & 26.8 & 29.3 & \textbf{40.4} \\
WMT23 & 5-shot, BLEURT & 65.2 & \textbf{72.0} & 59.9 & 65.3 & 71.1 \\
MGSM & 8-shot & 26.3 & 74.9 & 46.8 & 53.5 & \textbf{80.7} \\
\midrule
Average & & 39.0 & 59.6 & 46.1 & 50.6 & 64.5 \\
\bottomrule
\end{tabular}
}
\vspace{\baselineskip}
\caption{\label{tab:detailed_results} Detailed results on different tasks for PT and RLHFed models. We compare Gemma 2 and encoder-decoder models adapted via PrefixLM. Best results are in \textbf{bold}.}
\end{table*}

\paragraph{Pretraining objective matters: UL2 and PrefixLM show different characteristics.}

Previous study reported the superiority of UL2 over PrefixLM~\cite{tay2022ul2}, but PrefixLM in our study is enhanced with knowledge distillation, which often improves small models significantly. We compare these two objectives for the adaptation in Table \ref{tab:model_performance}.

We find that PrefixLM and UL2 have their own strengths. Specifically, UL2 delivers stronger contextual representations, outweighing PrefixLM on SuperGLUE across most model scales, resonating with previous findings~\cite{tay2022ul2}. In contrast, PrefixLM produces more powerful generative LLMs thanks to its generation nature and the knowledge distillation. It surpasses UL2 on PT and IT benchmarks in most cases. Particularly, it outperforms UL2 at 9B-2B on both PT and IT by up to 3.6, a significant margin. Since generative LLMs have become the mainstream, we base our following analysis on PrefixLM. We discuss our attempts to combine PrefixLM and UL2 in the next section.

\paragraph{Encoder-decoder LLMs outperform decoder-only LLMs especially after instruction tuning.}

Table \ref{tab:model_performance} also shows that the adapted encoder-decoder LLMs achieve comparable or slightly better pretraining performance than their decoder-only counterpart but with substantially improved instruction-tuning performance, echoing with the findings of~\citet{pmlr-v162-wang22u}. For example, the 9B-9B encoder-decoder LLM surpasses Gemma 2 9B by 1.4 and 4.9 on PT and IT, respectively. The performance gap further increases to 1.8 and 7.1 at 2B-2B scale. We notice that the adaption performs slightly worse at scales below 2B on PT, but the improvements on IT are still promising, e.g. 7.2 at XL-XL.

Regardless of PT or IT models, pretraining objectives, and model scales, encoder-decoder LLMs perform consistently better than decoder-only LLMs on SuperGLUE. This suggests that the contextual representation from encoder-decoder LLMs is often of higher quality, likely due to bidirectional self-attention. 

We need to highlight that the above analysis is based on the overall performance, which may not apply when it comes to a specific downstream task. As shown in Table \ref{tab:detailed_results}, there are some tasks favoring encoder-decoder models while others favoring decoder-only models especially for PT models. For example, after pretraining, Gemma 2 9B surpasses 9B-9B by 4.1 on ARC-C but underperforms it by 4.4 on Winogrande; while encoder-decoder LLM shows more consistent advantage after instruction tuning, 9B-9B still lags behind Gemma 2 9B by 0.9 on WMT23. This illustrates the complexity when evaluating LLM capability as well as the risk of reaching misleading conclusions when adopting biased evaluation tasks. We reduce such risk by selecting as diverse and broad tasks as possible for evaluation.

\begin{figure*}[t]
\centering
\small

\includegraphics[width=0.31\textwidth]{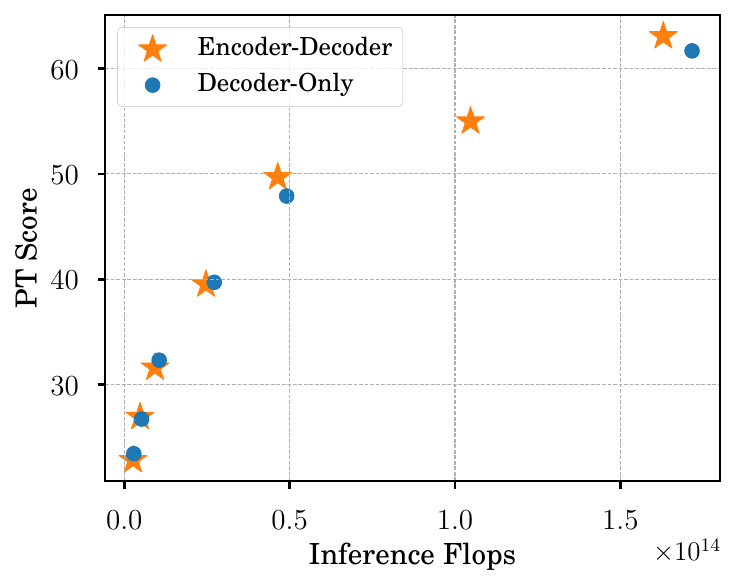}
\includegraphics[width=0.31\textwidth]{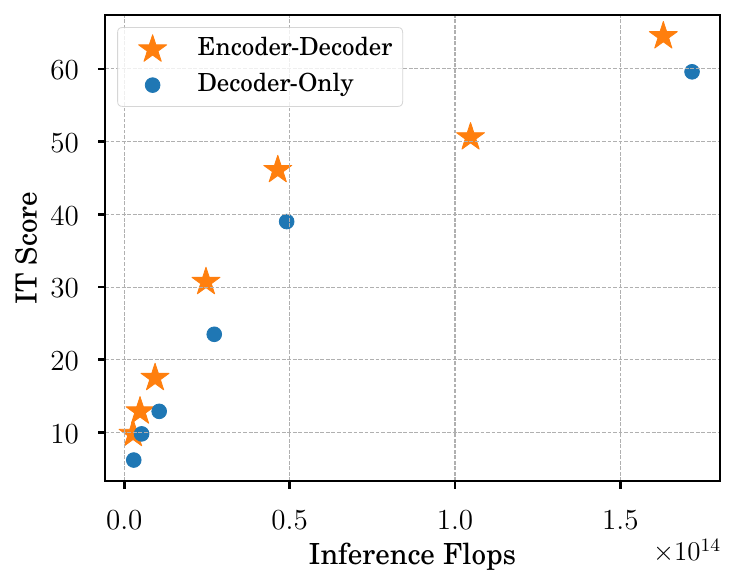}
\includegraphics[width=0.31\textwidth]{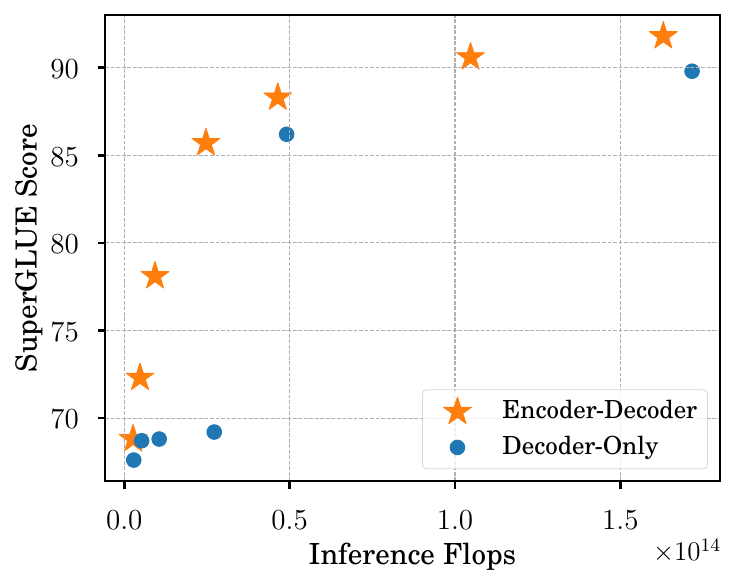}
\caption{\label{fig:performance_vs_flops} Comparisons of decoder-only LLMs with adapted encoder-decoder models under inference flops. We show PT, IT, and SuperGLUE performance. Inference flops is estimated with a sequence length of 4096-4096 and 8192 for encoder-decoder and decoder-only LLMs, respectively. Note the upper left corner marks the quality-efficiency frontier.}
\end{figure*}

\begin{figure}[t]
\centering
\small

\includegraphics[width=0.33\textwidth]{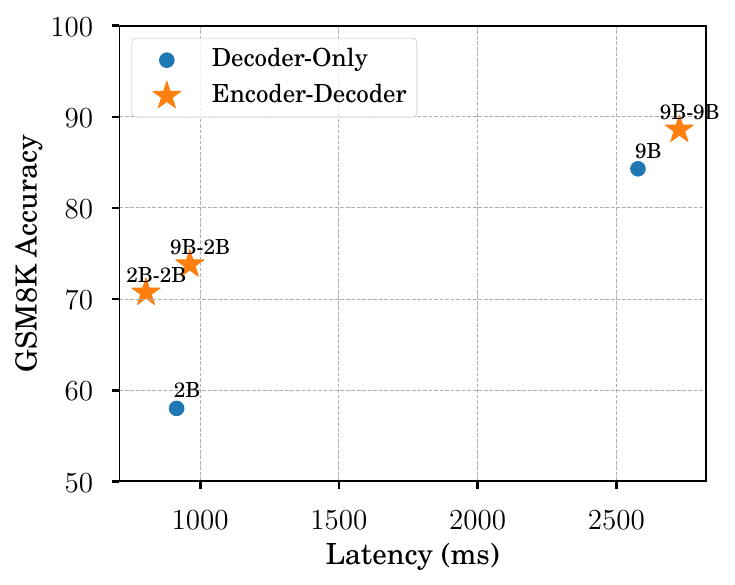}
\caption{\label{fig:latency_analysis} GSM8K performance as a function of latency for RLHFed models. Latency is estimated as milliseconds (ms) per query by answering 200 reasoning questions from GSM8K. Batch size of 1 is used.}
\end{figure}

\paragraph{Encoder-decoder LLMs balance quality and inference efficiency more effectively.}

We next analyze different models from the perspective of inference efficiency which becomes increasingly crucial for model deployment. Figure \ref{fig:performance_vs_flops} shows that balanced encoder-decoder LLMs have similar inference flops to their decoder-only counterparts, e.g. 2B-2B vs. Gemma 2 2B. As such, encoder-decoder models often dominate the quality-inference efficiency frontier across PT, IT, and SuperGLUE benchmarks.

We acknowledge that inference flops may not correlate well with actual running speed due to factors like inter-device communication, key-value caching, and autoregressive bottleneck. We then provide the latency results measured on GSM8K for 2B and 9B models in Figure \ref{fig:latency_analysis}, which further verified the above analysis. 9B-9B and 2B-2B show similar latency to Gemma 2 9B and 2B, respectively, but clearly better performance. In particular, 9B-2B, the one pairing large encoder and small decoder, shows similar latency to Gemma 2 2B but significantly better performance than 2B-2B. 

Together, these confirm that encoder-decoder adaptation indeed provides a more flexible way for balancing between quality and inference speed.

\section{Discussion}\label{sec:discussion}

\paragraph{Is the improvement after the adaptation simply due to the extra pretraining compute?}

Not really. We also tried to apply more pretraining compute to Gemma 2 2B by going through another 6 trillion tokens, which leads to a PT score of 48.57, still significantly below the encoder-decoder adaptation, 49.7. This indicates that the additional pretraining compute can't fully explain the improvements from the adaptation and we argue that the inductive bias of encoder-decoder modeling plays a crucial role.

\paragraph{Does cross-attention warmup matter for unbalanced encoder-decoder?}

Yes. Our preliminary experiments with 9B-2B and UL2 on 800B tokens show that the pretraining performance over Boolq and GSM8K reduces from 62.5 to 61.8 without the warmup. Besides, increasing warmup steps from 1K to 5K further reduces performance to 60.2. An adequate amount of warmup optimization is required to reach the optimal performance.

\paragraph{Can we switch from grouped-query attention to multi-head self attention for the encoder?}

Yes but with mixed results. Gemma 2 adopts grouped-query attention (GQA) to improve its decoding efficiency. However, unlike the decoder, the encoder can be fully parallelized during inference, making the use of multi-head attention (MHA) reasonable. We tried to expand GQA in Gemma 2 2B to MHA by replicating head parameters for the encoder self-attention. Under PrefixLM, this improves PT performance to 50.2 by 0.5 at 2B-2B but reduces IT performance to 43.5 by 2.9. We thus still stick to GQA when adapting Gemma 2 2B and 9B for the encoder.

\paragraph{Does bidirectional self-attention matter for the encoder?}

Yes. A crucial difference between encoder-decoder and decoder-only LLMs is the use of bidirectional self-attention. We also tested keeping the encoder self-attention causal at 2B-2B, which achieves a PT and IT score of 45.6 and 41.7, lagging behind its bidirectional counterpart substantially by 4.1 and 4.7, respectively. Note, the causal 2B-2B model surpasses Gemma 2 2B on IT by 2.7, although it performs worse on PT. This suggests that bidirectional self-attention contributes greatly to the success of our adaptation, but is not the only factor.

\begin{table}[t]
\centering
\centering
\small

\begin{tabular}{lrrrrrr}
\toprule

& \multicolumn{3}{c}{Adaptation} & \multicolumn{3}{c}{Scratch} \\
\cmidrule(lr){2-4} \cmidrule(lr){5-7}
& PT & IT & SG & PT & IT & SG \\
\midrule
S-S & 22.8 & 9.8 & 68.8 & \textbf{24.0} & \textbf{9.9} & \textbf{70.5} \\
B-B & 26.9 & \textbf{12.9} & 72.3 & \textbf{28.1} & 11.8 & \textbf{75.5} \\
L-L & \textbf{31.6} & \textbf{17.5} & 78.1 & 30.9 & 17.1 & \textbf{78.5} \\
XL-XL & \textbf{39.5} & \textbf{30.7} & \textbf{85.7} & 37.7 & 28.8 & 79.5 \\
2B-2B & \textbf{49.7} & \textbf{46.4} & \textbf{88.3} & 47.1 & 43.9 & 84.5 \\
\bottomrule
\end{tabular}

\caption{\label{tab:performance_scratch} Results for encoder-decoder models adapted with PrefixLM (\textit{Adaptation}) and pretrained from scratch (\textit{Scratch}). \textit{SG}: SuperGLUE score for SFTed models.}
\end{table}

\begin{figure*}[t]
\centering
\small

\includegraphics[width=0.31\textwidth]{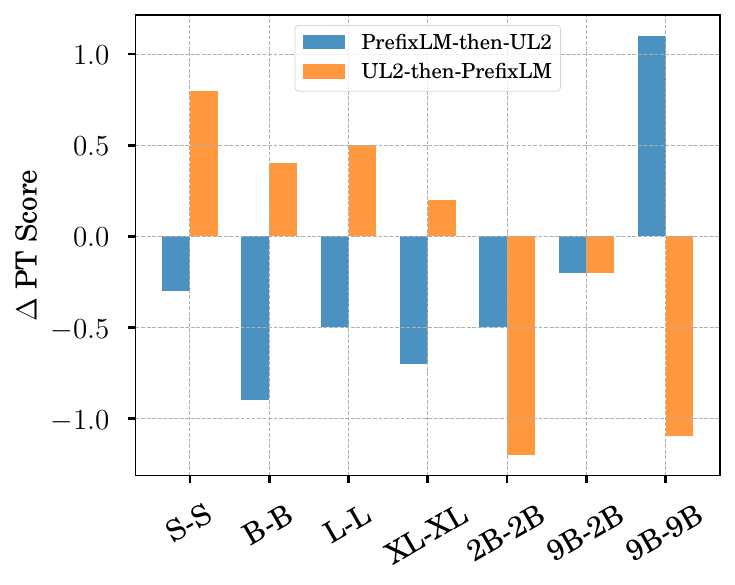}
\includegraphics[width=0.31\textwidth]{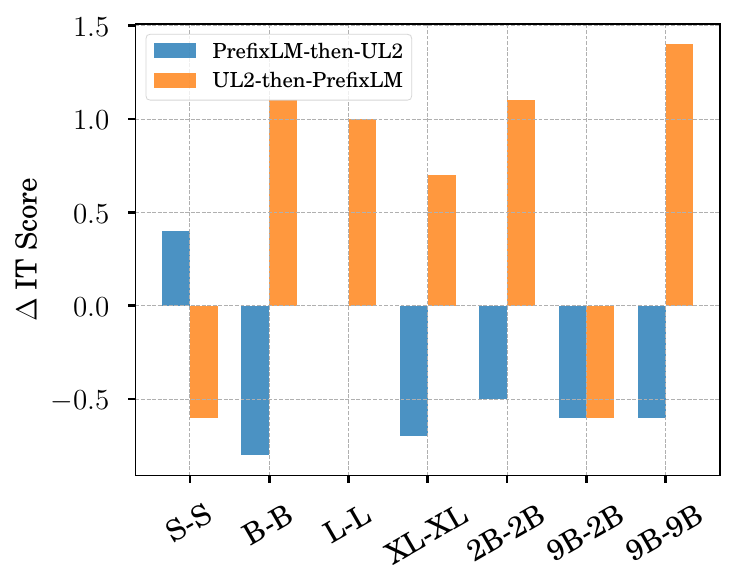}
\includegraphics[width=0.31\textwidth]{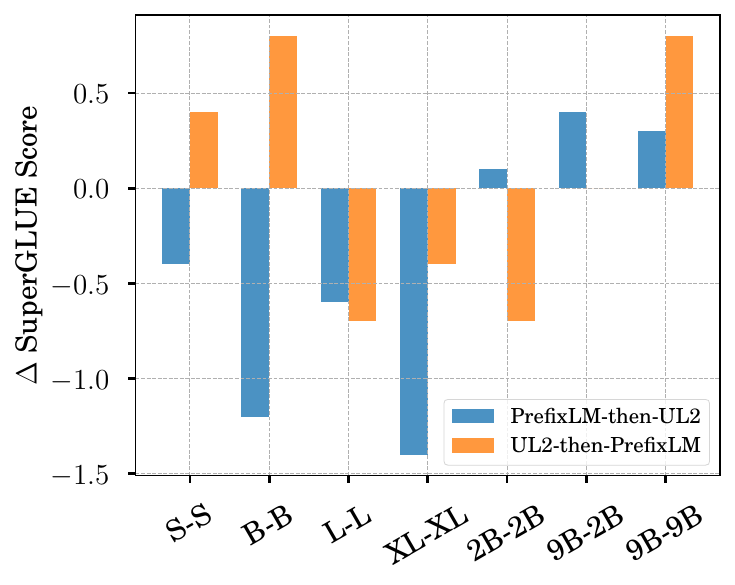}
\caption{\label{fig:ul2prefix_adapt_analysis} Quality change for the two-stage optimization. ``UL2-then-PrefixLM'': switch the training objective from UL2 to PrefixLM for the final 10\% tokens; ``PrefixLM-then-UL2'': similar but from PrefixLM to UL2.}
\end{figure*}

\begin{figure}[t]
\centering
\small

\includegraphics[width=0.33\textwidth]{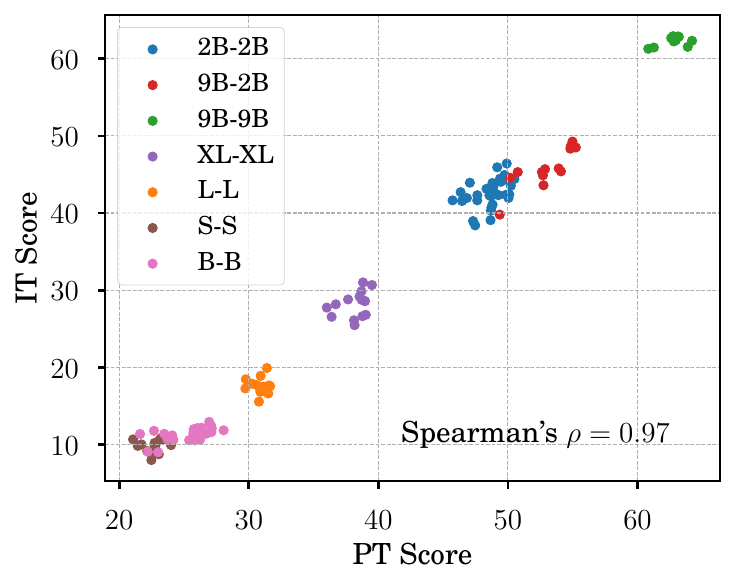}
\includegraphics[width=0.33\textwidth]{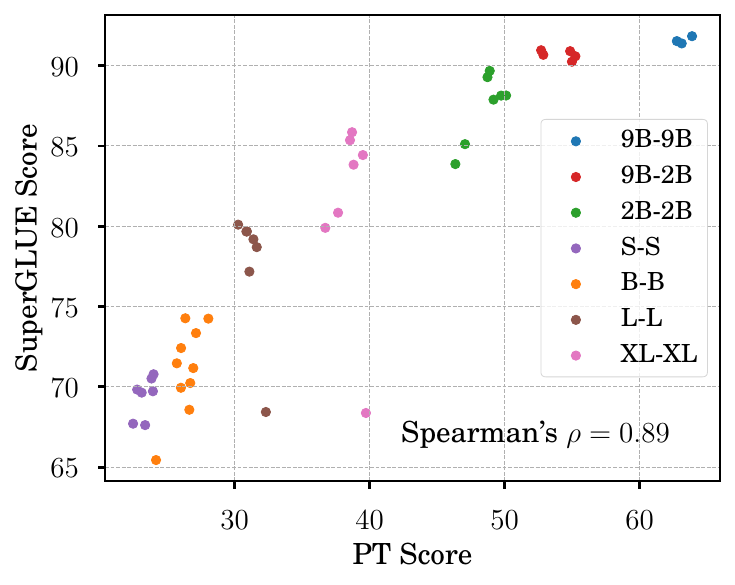}
\caption{\label{fig:coor_analysis} Correlation analysis between PT performance and its corresponding IT/SuperGLUE performance.}
\end{figure}

\paragraph{Would pretraining encoder-decoder LLMs from scratch yield better performance?}

Not really. Pretraining from scratch is a common method for developing new LLMs. We also pretrained encoder-decoder LLMs from scratch on 8 trillion tokens with PrefixLM. Table \ref{tab:performance_scratch} summarizes the results. Despite using more pretraining tokens, encoder-decoder LLMs pretrained from scratch only perform better at small scales, such as S-S and B-B, beyond which adaptation shows clear superiority. As such, adaptation is a more computationally efficient way of developing powerful encoder-decoder LLMs.

\paragraph{Is IT/SuperGLUE score predicable from PT score?}

Mixed. A general assumption in LLM development is that PT performance can be used as an indicator for downstream applications. We summarize all our ablations and put them in Figure \ref{fig:coor_analysis}. Over all data points and across all model sizes, the correlation is pretty strong: a Spearman's $\rho$ of 0.97 and 0.89 for IT vs. PT and SuperGLUE vs. PT, respectively. However, when considering data points within each model size separately, the averaged Spearman's $\rho$ reduces to 0.42 and 0.05, respectively and is not significant anymore.

In practice, we also noticed that PT checkpoints with weaker performance sometimes yield significantly better IT or SuperGLUE performance. When selecting PT checkpoints for a specific model size, it's better to also examine their IT performance apart from PT results to avoid some biases or overfitting.

\paragraph{Can we get the best of both worlds from PrefixLM and UL2?} This is non-trivial.
Our first attempt is to merge checkpoints trained from PrefixLM and UL2 with uniform weighting. Unfortunately, the merged model results in either similar or much worse performance. We argue that PrefixLM and UL2 lead to different training dynamics and converge to very different local minima. Directly merging their weights doesn't work right out of the box.

We next explore a two-stage optimization, where we first adapt with PrefixLM and then shift to UL2 for the last 10\% of training, and vice versa. Figure \ref{fig:ul2prefix_adapt_analysis} shows very mixed results. Switching from PrefixLM to UL2 generally hurts performance. In contrast, switching from UL2 to PrefixLM improves IT performance, but suffers from reduction in PT and SuperGLUE performance. 

Another direction is to jointly optimize the model on PrefixLM and UL2, which we leave for future work.

\section{Conclusion and Future Work}

In this paper, we presented methods for building powerful, general purpose encoder-decoder LLMs by adapting from pretrained decoder-only LLMs. Such adaptation offers high flexibility in leveraging different types/families of pretrained decoder-only models as well as combining different-sized models. Through extensive experiments based on Gemma 2, we demonstrated the feasibility and effectiveness of the adaptation: the adapted encoder-decoder LLMs outperform their decoder-only counterparts substantially after instruction tuning, dominating the quality-inference efficiency frontier. Besides, encoder-decoder LLMs also provide better contextual representations as evaluated on SuperGLUE.

We hope our findings inspire more researchers from academia and industry to revisit the encoder-decoder paradigm for LLM development. To facilitate the research, we will release the code and checkpoints at XXX (coming soon).

Our work still suffers from several limitations. Particularly, we only experimented with Gemma 2 models up to 9B, although the proposed approach could apply to other LLM families. In the future, we are interested in scaling the model size (e.g, to 27B), exploring other LLMs (such as LLaMA), examining more unbalanced setups, and testing the combination of dense and MoE LLMs. As mentioned above, we will also investigate better ways to leverage PrefixLM, knowledge distillation, and UL2. Extending our adapted encoder-decoder LLM to cross/multi-modality modeling (e.g., vision-language and speech-language) would be another intriguing direction. 

\section*{Acknowledgements}

We'd like to thank Enrique Alfonseca, Tris Warkentin, Xiaodan Song, Sugato Basu, Inderjit Dhillon, Alexander Grushetsky, Pandu Nayak, Ramakrishnan Srikant, and Slav Petrov for their constructive feedback on the manuscript. We are grateful to Srinivasan Venkatachary for supporting this project.


\nocite{langley00}

\bibliography{paper}
\bibliographystyle{icml2025}

\appendix
\onecolumn


\end{document}